# Multi-Product Inventory Optimization using Uniform Crossover Genetic Algorithm

**S.Narmadha**
Assistant Professor
Department of Computer Science and Engineering
Park College of Engineering and Tekhnology
Coimbatore – 641659, Tamilnadu, India

**Dr.V.Selladurai**
Professor and Head
Department of Mechanical Engineering
Coimbatore Institute of Technology
Coimbatore – 641014, Tamilnadu, India

**G.Sathish**
Research Scholar
Department of Computer Science and Engineering
Anna University – Coimbatore, Tamilnadu, India

*Abstract -* **Inventory management is considered to be an important field in Supply Chain Management because the cost of inventories in a supply chain accounts for about 30% of the value of the product. The service provided to the customer eventually gets enhanced once the efficient and effective management of inventory is carried out all through the supply chain. The precise estimation of optimal inventory is essential since shortage of inventory yields to lost sales, while excess of inventory may result in pointless storage costs. Thus the determination of the inventory to be held at various levels in a supply chain becomes inevitable so as to ensure minimal cost for the supply chain. The minimization of the total supply chain cost can only be achieved when optimization of the base stock level is carried out at each member of the supply chain. This paper deals with the problem of determination of base-stock levels in a ten member serial supply chain with multiple products produced by factories using Uniform Crossover Genetic Algorithms. The complexity of the problem increases when more distribution centers and agents and multiple products were involved. These considerations leading to very complex inventory management process has been resolved in this work.**

**Keywords: Supply Chain Management, Inventory Optimization, Base Stock, Uniform Crossover, Genetic Algorithm (GA), Supply Chain Cost**

## I. INTRODUCTION

Supply Chain Management (SCM) is an efficient management of the complete end to end process, starting from the design of the product or service to the time when it has been sold, consumed and finally gotten rid of by the consumer. This complete process includes product design, procurement, planning and forecasting, production, distribution, fulfillment and after sales supports. A company's competitiveness in the global economy can be increased only with the aid of effective SCM. This involves complex strategic, tactical and operational decisions that often require an in-depth understanding of industry-specific issues, which ranges from network design to production sourcing and from production planning and inventory management to scheduling [1].

The inventory management problem is one of maintaining an adequate supply of some item to meet an expected pattern of demand, while striking a reasonable balance between the cost of holding the items in inventory and the penalty (loss of sales and goodwill, say) of running out. The item may be a commodity sold by a store; it may be spare machine parts in a factory; it may be railway wagons; it may be cash in the bank to meet the customers' demand. It is indeed surprising to find that a very wide variety of seemingly different problems can be mathematically formulated as an inventory-control problem. There are, of course, several different models of inventory systems. There are three types of expenses associated with inventory systems. The relative importance of these will depend on the specific system. They are: (i) administrative cost of placing an order, called reorder cost or set cost; (ii) cost of maintaining an inventory, called inventory holding cost a carrying cost, which includes storage charge, interest, insurance, etc., a (iii) shortage cost is a loss of profit, goodwill, etc., when run out of stock. All the above should be optimized for efficient supply chain management.

### A. Inventory Control in Supply Chain Management

It has been stated by several people that the focus point of supply chain management is inventories and inventory control. To transfer their focus from scheming logistical costs to investigate supply chains [2] few food manufacturers and grocers formed Efficient Consumer Response in the year 1992. The major competitive factor for companies focused on value creation for end consumers is the customer service. In general, firms hold inventory for two major reasons, to lessen costs and to improve customer service. The inspiration for each varies as firms stabilize the problem of having too much inventory (which can direct to high costs) versus having very small inventory (which can direct to lost sales) [3].

Supply chain management leads to cost savings, mainly in the course of lessening in inventory. Inventory costs have got reduced by about 60% from 1982, whereas transportation costs have fallen by 20% [4]. These cost savings have led many people to follow inventory-reduction strategies in the supply chain. To deal with inventory, firms make use of one of three common approaches. First of all, the majority of retailers make use of an inventory control approach, monitoring inventory levels by item. The second thing is, manufacturers are typically more concerned with production scheduling and use flow management to deal with inventories.

Third, numerous firms (for the majority part those handling raw materials or in extractive industries) do not keenly deal with inventory [5].

The inventory management is influenced by the nature of demand, depending on whether demand is derived or independent. Independent demand comes up from demand for an end product. End products are found all through the supply chain. By definition, a self-governing demand is uncertain, meaning that extra units or safety stock must be accepted to guard against stock outs. While managing uncertainty, the objective should be to minimize the inventory levels and also meet customer expectation. Supply chain coordination can reduce the ambiguity of intermediate product demand, in that way reducing inventory costs [3, 6].

Since Ford Harris' renowned Economic Order Quantity (EOQ) model was first proposed in 1913, the inventory control has been rewarded immense awareness for a long time because of its significance in the cost control. To lessen the total expected inventory costs per unit time while satisfying the customer demand on time [7] is one of the major objectives. Inventory control for large-scale supply chains is well recognized [8-10] as an essential problem with several applications together with manufacturing systems, logistics systems, communication networks, and transportation systems [11]. It is essential to locate the apt mechanism for coordinating the inventory processes that are controlled by independent partners, in order to find out the right ordering quantity and inventory level amid partners in the chain. For example, the manufacturer make use of the periodic review and lot sizing policy to manage its inventory and the retailer employs the periodic review with target stock level to control its inventory and more [12].

### B. Inventory Optimization in Supply Chain Management

The effective management of the supply chain has become unavoidable these days due to high expectation in customer service levels [13]. The supply chain cost was immensely influenced by the overload or shortage of inventories. Thus inventory optimization has transpired into one of the most important topics as far as supply chain management is considered [14-16].

To exploit economies of scale and order in large lots, the important issues in supply chain is to optimize the inventory level by considering various costs in maintaining a high service level towards the customer. Since, the cost of capital tied up in inventory is more, the inventory decision in the supply chain should be coordinated without disturbing the service level. The coordination of inventory decision within an entity is viable, but not between the entities. So the integration of the entities to centralize the inventory control is needed.

Inventory Optimization [IO] application organizes the latest techniques and technologies, thereby assisting the improved inventory visibility, the enhancement of inventory control and its management across an extended supply network. Some of the design objectives of inventory optimization are to optimize inventory strategies, thereby enhancing customer service, reducing lead times and costs and meeting market demand [14-16]. The design and management of the storage policies and procedures for raw materials, work-in-process inventories, and typically, final products are illustrated by the inventory control. The costs and lead times can be reduced and the responsiveness to the changing customer demands can be significantly improved and subsequently inventory can be optimized by the effective handling of the supply chain [17].

There are several reasons for manufacturers' increasing focus on optimizing inventory by applying the latest tools and techniques for inventory control. Traditionally, competitive pressure has always driven manufacturers to seek enhanced capabilities to reduce inventory levels; to enhance service levels and supply availability; and to establish the right product inventory mix and level in each geography and channel. A key driver of the renewed focus on inventory lies in the recognition that traditional techniques are failing to reign in inventories in the wake of increased supply chain complexity. This complexity is characterized by increased uncertainty. Demand is more volatile and therefore less predictable. This is true not only for aggregate demand but for forecasting splits and volumes across channels and markets. Traditionally three strategies have been employed by manufacturers to address uncertainty; a) increase inventory levels to hedge against uncertainty; b) develop supply chain flexibility to be more responsive to uncertainty; c) improve forecast accuracy so that less uncertainty propagates to the manufacturing floor. Inventory optimization techniques and technologies map to the flexibility and accuracy strategies. [18].

Inventory Optimization characterizes the supply network uncertainty present in a variety of specific steps or links in manufacturing and distribution processes. Advanced mathematical models are then solved to identify optimal inventory policies, stocking locations, or quantities. The uncertainty addressed by IO include: demand uncertainty, cycle time variability and replenishment lead time variability.[18] Efficient management of the supply chain, i.e. the reduction of the costs and lead times and vastly enhanced responsiveness to the changing customer demands lead to an optimized inventory.

## II. RELATED WORKS - REVIEW

### A. Review of Base-Stock based Inventory Control Models

The inventory control problem for a single class assembly network which operates under a modified echelon base-stock policy was studied by Vecchio and Paschalidis [19]. An approach to find close-to-optimal echelon stock levels that minimize inventory costs while guaranteeing stockout probabilities to stay below some predefined levels was developed by them. They reduced the safety stock selection to a deterministic nonlinear optimization problem on the basis of the large deviations techniques. In addition, they analyzed as to how a supplier can interact with a buyer to reach a mutually beneficial mode of operations, using their inventory control approach. Their interaction takes the form of a supply contract by which explicit QoS guarantees is enforced. The applications in a distributed fashion with neither the supplier nor the buyer revealing their corresponding cost structures applying the joint optimization algorithm was proposed by the authors.

A model of supply chain consisting of n production facilities in tandem and producing a single product class was considered by Ioannis CH. Paschalidis et al. [20]. The finished goods inventory maintained in front of the most downstream facility is used to meet the external demand while backlogging of unsatisfied demand was performed. The facility at stage 1 produced if inventory has fallen below a certain level $w_i$ and idles otherwise on the basis of a base-stock production policy adopted at each stage of the supply chain. In order to minimize expected inventory costs at all stages subject to maintaining the stock out probability at stage1 below a prescribed level, they necessitated the optimization of the hedging vector W= $(w_1,....,w_n)$. They made assumptions on demand and production processes that included auto correlated stochastic processes, which were relatively general modeling. They have combined analytical (Large derivations) and sample path based (perturbation analysis) techniques to solve the stochastic optimization problem. The existence of a natural synergy between those two approaches has been demonstrated.

An attempt was made to optimize the inventory (i.e. base-stock) levels of a single product at different members in a serial supply chain with the objective of minimizing the Total Supply Chain Cost (TSCC), by Sudhir Ryan Daniel and Chandrasekharan Rajendran [21] and P. Radhakrishnan et al. [22]. The performance measure considered, which is a good representation of the system-wide total cost is the TSCC. In order to optimize the base-stock levels, a genetic algorithm (GA) has been proposed. To analyze the performance of the supply chain (operating with deterministic and stochastic replenishment lead times) for the base-stock levels that are generated by the proposed GA and other solution procedures considered in this study, different supply chain settings are simulated. They demonstrated that their proposed GA required significantly less computing effort to perform very well in terms of yielding solutions that are not significantly different from the optimal solutions (obtained through complete enumeration of solution space).

A beneficial industry case applying Genetic Algorithms (GA) has been proposed by K.Wang and Y.Wang [23]. The case has made use of GAs for the optimization of the total cost of multiple sourcing supply chain system. The system has been exemplified by a multiple sourcing model with stochastic demand. A mathematical model has been implemented to portray the stochastic inventory with the many to many demand and transportation parameters as well as price uncertainty factors A genetic algorithm which has been approved by Lo [24] deals with the production-inventory problem with backlog in the real situations, with time-varied demand and imperfect production due to the defects in production disruption with exponential distribution. Besides optimizing the number of production cycles to generate a (R, Q) inventory policy, an aggregative production plan can also be produced to minimize the total inventory cost on the basis of reproduction interval searching in a given time horizon.

The inventory levels across supply chain members were obtained with the aid of a search routine.

### B. Review of Optimization based Inventory Control Models

Sukran Kadipasaoglu et al. [1] provided a study on the market characteristics and competitive priorities, manufacturing environment, logistics and distribution activities, and supply chain planning and control activities for polymer manufacturers. They have described polymer distribution network optimization, production/distribution planning, production scheduling, demand management, available-to-promise, and inventory planning activities pertaining to supply chain planning and control. Besides, they illustrated the applications existing in a commercial DSS that support these activities. They have as well described about diverse issues that continue to confront supply chain managers in polymer manufacturing. It encompasses forecasting for the huge number of product–customer combinations, identification of safety stock requirements, administering production schedule changes, business process management throughout DSS implementation and data mapping for decision support. Their research contributes to the supply chain literature by proffering a suitable context for investigating supply chain-related issues. Through discussion and characterization of the polymer supply chain, they recognized the specific issues of concern to potential researchers and to supply chain professionals.

The effect of product variety on supply-chain performance, which is measured in terms of expected lead time and expected cost at the retailers, was analyzed by Ulrich W. Thonemann and James R. Bradley [25]. They took a supply chain with a single manufacturer and multiple retailers into account. If setup times are significant, the effect of product variety on cost where the cost increases proportionally to the square root of product variety is substantially greater than that suggested by the risk-pooling literature for perfectly flexible manufacturing processes. An illustration that underestimates the cost of product variety, leads companies to offer product variety that is greater than optimal was made as well. In conclusion, they illustrated that by reducing the setup time, the unit manufacturing time, the number of retailers, or the demand rate the supply-chain performance can be managed. The fact that complex mathematical approaches are often not applied in practice was recognized by the authors. Nevertheless, practitioners who used the simple models to estimate the effect of their decisions often appreciated these models.

The inventory and supply chain managers are mainly concerned holding of the excess stock levels and hence the increase in the holding cost. Meanwhile, there is possibility for the shortage of products. For the shortage of each product there will be a shortage cost. Holding excess stock levels as well as the occurrence of shortage for products lead to the increase in the supply chain cost. The factory may manufacture any number of products, each supply chain member may consume a few or all the products and each product is manufactured using a number of raw materials sourced from many suppliers. All these factors pose additional challenge in extracting the exact product and the stock levels that influence the supply chain cost heavily.

Many well-known algorithmic advances in optimization have been made, but it turns out that most have not had the expected impact on the decisions for designing and optimizing supply chain related problems. For example, some optimization techniques are of little use because they are not well suited to solve complex real logistics problems in the short time needed to make decisions. Also some techniques are highly problem-dependent and need high expertise. This adds difficulties in the implementations of the decision support systems which contradicts the tendency to fast implementation in a rapidly changing world. IO techniques need to determine a globally optimal placement of inventory, considering its cost at each stage in the supply chain and all the service level targets and replenishment lead times that constraint each inventory location about the estimation of the exact amount of inventory at each point in the supply chain free of excesses and shortages although the total supply chain cost is minimized. Owing to the fact that shortage of inventory yields to lost sales, whereas excess of inventory may result in pointless storage costs, the precise estimation of optimal inventory is indispensable [26]. In other words, there is a cost involved in manufacturing any product in the factory as well as in holding any product in the distribution center and agent shop. More the products manufactured or held, higher will be the holding cost. Along with this, low lead time results in

## III. OBJECTIVES

The supply chain cost can be minimized by maintaining optimal stock levels in each supply chain member. There is a necessity of determining the inventory to be held at different stages in a supply chain that will minimize the total supply chain cost i.e., minimizing holding and shortage cost**.** The inventory control for more number of products along with different levels of supply chain is a complex task. The approach aims to make use of the meta heuristic algorithms like Genetic algorithm for the prediction of the optimal stock levels to be maintained, so as to minimize the total supply chain inventory cost, comprising holding and shortage costs at all members of the supply chain. The genetic algorithm is proposed that considers all these factors that are mentioned hitherto such that the analysis paves the way for minimizing the supply chain cost by maintaining optimal stock levels in each supply chain member.

### A. Genetic Algorithm

Genetic algorithm is a randomized search methodology having its roots in the natural selection process. Initially the neighborhood search operators (crossover and mutation) are applied to the preliminary set of solutions to acquire generation of new solutions. Solutions are chosen randomly from the existing set of solutions where the selection probability and the solution's objective function value are proportional to each other and eventually the aforesaid operators are applied on the chosen solutions. Genetic algorithms have aided in the successful implementation of solutions for a wide variety of combinatorial problems.

The robustness of the Genetic algorithms as search techniques have been theoretically and empirically proved [27]. The artificial individual is the basic element of a GA. An artificial individual consists of a chromosome and a fitness value, similar to a natural individual. The individual's likelihood for survival and mating is determined by the fitness function [28]. In accordance with the Darwin's principle, individuals superior to their competitors, are more likely to promote their genes to the next generations. In accordance with this concept, in Genetic Algorithms, a set of encoded parameters are mapped into a potential solution, named chromosome, to the optimization problem [29]. The population of candidate solutions is obtained through the process of selection, recombination, and mutation performed in an iterative manner. [30].

Chromosomes refer to the random population of encoded candidate solutions with which the Genetic algorithms initiate with. [27]. Then the set (called a population) of possible solutions (called chromosomes) are generated [31]. A function assigns a degree of fitness to each chromosome in every generation in order to use the best individual during the evolutionary process [32]. In accordance to the objective, the fitness function evaluates the individuals [30]. Each chromosome is evaluated using a fitness function and a fitness value is assigned. Then, three different operators-selection, crossover and mutation- are applied to update the population. A generation refers to an iteration of these three operators [33]. The promising areas of the search space are focused in the selection step. The selection process typically keeps solutions with high fitness values in the population and rejects individuals of low quality [30]. Hence, this provides a means for the chromosomes with better fitness to form the mating pool (MP) [31]. After the process of Selection, the Crossover is performed.

### B. Uniform Crossover

In the crossover operation, two new children are formed by exchanging the genetic information between two parent chromosomes. Multipoint crossover defines crossover points as places between loci where an individual can be split. Uniform crossover generalizes this scheme to make every locus a potential crossover point. A crossover mask, the same length as the individual structure is created at random and the parity of the bits in the mask indicate which parent will supply the offspring with which bits. This method is identical to discrete recombination.

Consider the following two individuals with 11 binary variables each:

| Individual 1 | 0 1 1 1 0 0 1 1 0 1 0 |
|---|---|
| Individual 2 | 1 0 1 0 1 1 0 0 1 0 1 |

For each variable the parent who contributes its variable to the offspring is chosen randomly with equal probability. Here, the offspring 1 is produced by taking the bit from parent 1 if the corresponding mask bit is 1 or the bit from parent 2 if the corresponding mask bit is 0. Offspring 2 is created using the inverse of the mask, usually.

| Sample 1 | 0 1 1 0 0 0 1 1 0 1 0 |
|---|---|
| Sample 2 | 1 0 0 1 1 1 0 0 1 0 1 |

After crossover the new individuals are created:
offspring 1    1 1 1 0 1 1 1 1 1 1
offspring 2    0 0 1 1 0 0 0 0 0 0

Uniform crossover has been claimed to reduce the bias associated with the length of the binary representation used and the particular coding for a given parameter set. This helps to overcome the bias in single-point crossover towards short substrings without requiring precise understanding of the significance of the individual bits in the individual's representation. How uniform crossover may be parameterized by applying a probability to the swapping of bits was demonstrated by William M. Spears et al.[34].

This extra parameter can be used to control the amount of disruption during recombination without introducing a bias towards the length of the representation used. The chromosome cloning takes place when a pair of chromosomes does not crossover, thus creating off springs that are exact copies of each parent [28].

The ultimate step in each generation is the mutation of individuals through the alteration of parts of their genes [26]. Mutation alters a minute portion of a chromosome and thus institutes variability into the population of the subsequent generation [27]. Mutation, a rarity in nature, denotes the alteration in the gene and assists us in avoiding loss of genetic diversity [26]. Its chief intent is to ensure that the search algorithm is not bound on a local optimum [28].

## IV. INVENTORY OPTIMIZATION ANALYSIS USING UNIFORM CROSSOVER GENETIC ALGORITHM

The proposed method uses the Genetic Algorithm with Uniform Crossover to study the stock level that needs essential inventory control. This is the pre-requisite information that will make any kind of inventory control effective. In practice, the supply chain is of length $n$, means having $n$ number of members in supply chain such as factory, distribution centers, suppliers, retailers and so on. The exemplary supply chain taken for the implementation of the proposed method is shown in Fig. 1.

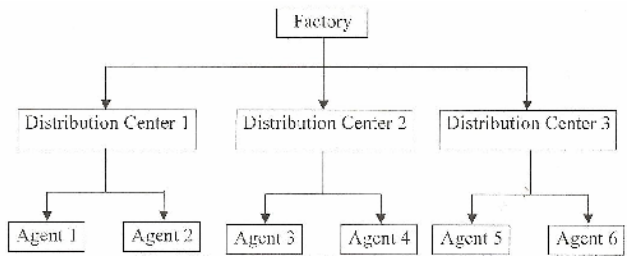

**Fig. 1 Three Stage - 10 Member Supply Chain**

Fig. 1 exhibits that a factory is the parent of the chain and it is having three distribution centers Distribution Center 1, Distribution Center 2 and Distribution Center 3. Each distribution center further comprises of several agents but as stated in the example case, each Distribution center is having two agents. So, in aggregate there are six agents, Agent 1 and Agent 2 for Distribution Center 1, Agent 3 and Agent 4 for Distribution Center 2 and Agent 5 and Agent 6 for Distribution Center 3. The factory manufactures different products that would be supplied to the distribution centers. From the distribution center, the stocks will be moved to the corresponding agents.

To make the inventory control effective, the most primary objective is to predict where, why and how much of the control is required which is made through the proposed GAUX methodology. The proposed methodology is aimed at determining the specific product that needs to be concentrated on and the amount of stock levels of the product to be maintained by the different members of the supply chain. The methodology also analyses whether the stock level of the particular product needs to be in abundance, in order to avoid shortage of the product or needs to be held minimal in order to minimize the holding cost.

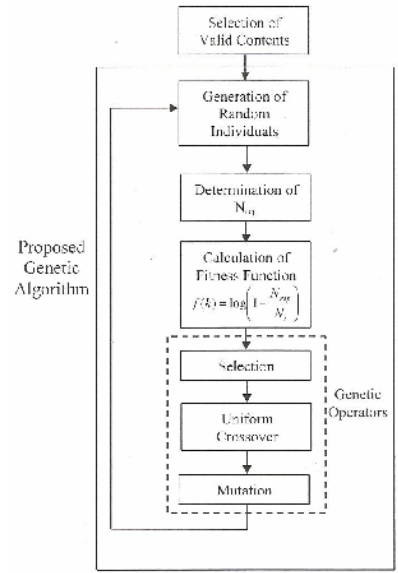

**Fig. 2 Genetic Algorithm steps for the proposed inventory management analysis**

The methodology as shown in Fig. 2 would analyze the past records very effectively and thus facilitate efficient inventory management with the aid of Genetic Algorithm. The analysis is initiated by the selection of valid records. The validation of records is done over the records of past periods. The stock levels at the different supply chain members are held in the dataset for different products, namely P1, P2, P3, P4, P5, P6, P7, etc. Seven products have been considered for the analysis, hence the stock levels for the seven products at each member of the chain throughout the past period are considered as data set as shown in the Table 1. For the valid record set selection, records having nil values are neglected and the records having positive or negative values are selected for the analysis. This can be done by means of clustering algorithms, extraction algorithms or by any of the data mining functions. Hence the extraction function results in data sets having either positive or negative values. The record set having positive values represents excess stock levels and the negative values represent shortage level of a particular product at a particular member of the supply chain. Then the data set is subjected to Genetic Algorithm and the various steps performed in the genetic algorithm are discussed below.

### A. Generation of Individuals

The randomly generated initial chromosome is created by having the stock levels within the lower limit and the upper limit for all the contributors of the supply chain, factory and the distribution centers. As known, chromosome is constituted by genes which defines the length of the chromosomes. The stock level of each member of the chromosome is referred as gene of the chromosome. Hence for *n* length supply chain, the chromosome length is also *n* . Since a 10 member supply chain is used for illustration, the length of the chromosome *n* is 10, i.e. 10 genes. And the chromosome representation is pictured in Fig. 3. Each gene of the chromosome is representing the amount of stock that is in excess or in shortage at the respective members of the supply chain.

| P3 | 7000 | -200 | -600 | -500 | 450 | -350 | 800 | -400 | 700 | -600 |
|----|------|------|------|------|-----|------|-----|------|-----|------|

| P2 | 5000 | 400 | -800 | 500 | 445 | 315 | -820 | 405 | -150 | 100 |
|----|------|-----|------|-----|-----|-----|------|-----|------|-----|

**Fig. 3 Random individual generated for the genetic operation**

These kinds of chromosomes are generated for the genetic operation. Initially, only two chromosomes will be generated and from the next generation a single random chromosome value will be generated. The chromosomes thus generated is then applied to find its number of occurrences in the database content by using a Select count ( ) function. The function will give the number of occurrences/ repetitions of the particular amount of stock level for the ten members $N_{rep}$ that are going to be used further in the fitness function.

### B. Evaluation of Fitness function

A specific kind of objective function that enumerates the optimality of a solution in a genetic algorithm in order to rank certain chromosome against all the other chromosomes is known as Fitness function. Optimal chromosomes, or at least chromosomes which are near optimal, are permitted to breed and merge their datasets through one of the several techniques available in order to produce a new generation that will be better than the ones considered so far.

The fitness function is given by:

$$f(k) = \log\left(1 - \frac{N_{rep}}{N_t}\right). \quad k = 1,2,3 \cdots\cdots, m \qquad (1)$$

where,

$N_{rep}$ is the number of repetitions of records of similar stock levels that occurs throughout the period;

$N_t$ is the total number of records of inventory values obtained after clustering;

*m* is the total number of chromosomes for which the fitness function is calculated.

In the fitness function, the ratio ($N_{rep}$ / $N_t$) plays the role of finding the probability of occurrence of a particular record of inventory values; and log [1- ($N_{rep}$ / $N_t$)] will ensure minimum value corresponding to the maximum probability; So, the fitness function is structured to retain the minimum value corresponding to the various chromosomes being evaluated iteration after iteration and this in turn ensures that the fitness function evolution is towards optimization.

### C. Selection

The selection operation is the initial genetic operation which is responsible for the selection of the fittest chromosome for further genetic operations. The fitness function is carried out for each chromosome and the chromosomes are sorted on the basis of the result of the fitness function and ranked. The chromosome generating value as minimum as possible will be selected by the fitness function and will be subjected further to the genetic operations, crossover and mutation.

### D. Uniform Crossover

Among the numerous crossover operators in practice, a uniform crossover is chosen in this proposed method for its advantages over the other forms. Uniform crossover is global and less biased when compared to that of standard and one point crossover. Uniform crossover does not select a set of crossover points. It simply considers each bit position of the two parents, and swaps the two bits with a probability of 50%. With large search spaces, a GA using uniform crossover outperforms a GA using one point crossover, which in turn outperforms a GA using two point crossover [35-36]. From the matting pool, two chromosomes are subjected for the uniform crossover. The chromosomes initially selected and after undergoing uniform crossover operation performed in this analysis is pictured in Fig. 4. As soon as the crossover operation is completed, the genes of the two chromosomes present get interchanged.

*Before Crossover*

| P3 | 7000 | -200 | -600 | -500 | 450 | -350 | 800 | -400 | 700 | -600 |
|----|------|------|------|------|-----|------|-----|------|-----|------|

| P2 | 5000 | 400 | -800 | 500 | 445 | 315 | -820 | 405 | -150 | 100 |
|----|------|-----|------|-----|-----|-----|------|-----|------|-----|

*After Crossover*

| P3 | 7000 | -150 | -820 | -400 | 315 | -350 | -750 | 405 | 500 | 600 |
|----|------|------|------|------|-----|------|------|-----|-----|-----|

| P2 | 5000 | 450 | 700 | -600 | 350 | -200 | 800 | 500 | -100 | 150 |
|----|------|-----|-----|------|-----|------|-----|-----|------|-----|

**Fig. 4 Chromosomes after uniform crossover operation**

### E. Mutation

The crossover operation is succeeded by the final stage of genetic operation known as Mutation. In the mutation, a new chromosome is obtained. This chromosome is totally new from the parent chromosome. The concept behind this is the child chromosome thus obtained will be fitter than the parent chromosome. The performance of mutation operation is shown in Fig. 5.

Before Mutation

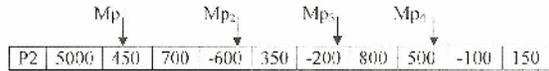

After Mutation

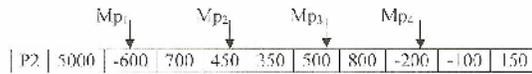

**Fig. 5 Chromosome subjected to mutation operation**

Four mutation points are chosen as shown in Fig. 5. The mutation is done on the particular gene present at the Mutation points. This pointing of gene is done randomly. Hence, the four mutation points may point any of the ten genes.

The process explained so far will be repeated along with the new chromosome obtained from the previous process. In other words, at the end of each of the iteration, a best chromosome will be obtained. This will be included with the newly generated random chromosome for the next iteration. When the number of iterations is increased then the obtained solution moves very closer to the accurate solution. More the number of iterations the more accurate the optimal solution will be. Eventually with the help of the Genetic algorithm, the best stock level to be maintained in the members of the supply chain could be predicted from the past records, so that the loss due to the holding of excess stock level and shortage level can be reduced leading to an optimal inventory solution.

## V. EXPERIMENTAL RESULTS

The approach suggested for the optimization of inventory level and thereby an efficient supply chain management has been implemented in the platform of LabVIEW 2009. The database consists of the records of stock levels held by each member of the supply chain for every period. For implementation, seven different products in circulation with the ten member supply chain network have been considered. A sample set of data from a large database used in the implementation is given in Table 1.

**Table 1. A Sample Dataset Constituted by the Product Identification along with its Stock Levels in Each Member of the Supply Chain**

| PI | F1 | DC1 | DC2 | DC3 | A1 | A2 | A3 | A4 | A5 | A6 |
|---|---|---|---|---|---|---|---|---|---|---|
| 7 | -371 | -736 | -299 | 634 | 448 | 756 | 340 | -736 | -778 | 863 |
| 5 | -407 | 379 | -981 | -864 | -391 | 999 | -196 | 307 | -171 | -529 |
| 2 | -146 | -604 | 443 | 746 | -561 | -734 | 445 | 424 | -891 | -824 |
| 4 | -962 | -524 | -685 | -254 | 205 | 446 | -469 | 108 | 346 | 840 |
| 3 | -834 | 266 | 969 | 965 | 735 | 244 | -752 | 133 | -554 | -939 |
| 3 | -449 | -282 | 577 | -926 | -414 | -200 | -743 | 850 | 196 | 851 |
| 4 | 540 | -830 | -835 | 882 | -379 | 768 | -635 | -112 | 539 | 107 |
| 3 | -778 | -313 | 629 | -690 | 824 | -927 | 850 | 307 | -171 | -529 |
| 2 | 351 | 293 | 328 | -732 | 357 | -566 | 685 | 424 | -891 | -824 |
| 1 | 500 | 108 | 490 | -345 | -236 | 108 | -931 | -260 | -144 | 162 |
| 5 | 844 | -728 | 286 | 740 | 686 | -421 | 424 | -792 | -927 | -879 |
| 4 | -321 | 902 | -450 | -260 | -144 | 162 | 238 | 307 | -171 | -529 |
| 3 | 775 | -394 | -520 | -792 | -927 | -879 | -507 | 424 | -891 | -824 |
| 4 | 794 | 932 | -584 | 307 | -171 | -529 | -503 | 108 | 346 | 840 |
| 2 | -122 | -686 | -620 | 424 | -891 | -824 | 941 | 133 | -554 | -939 |
| 6 | 235 | 464 | 401 | 108 | 346 | 840 | -934 | 464 | 401 | 108 |
| 5 | 218 | -848 | 836 | 133 | -554 | -939 | -834 | -848 | 836 | 133 |
| 4 | 489 | 409 | 148 | 850 | 196 | 851 | -495 | -144 | 162 | 238 |
| 3 | -422 | 638 | 676 | -112 | 539 | 107 | -440 | -927 | -879 | -507 |
| 5 | 893 | 520 | -423 | -736 | -778 | 863 | -335 | 676 | -112 | 539 |

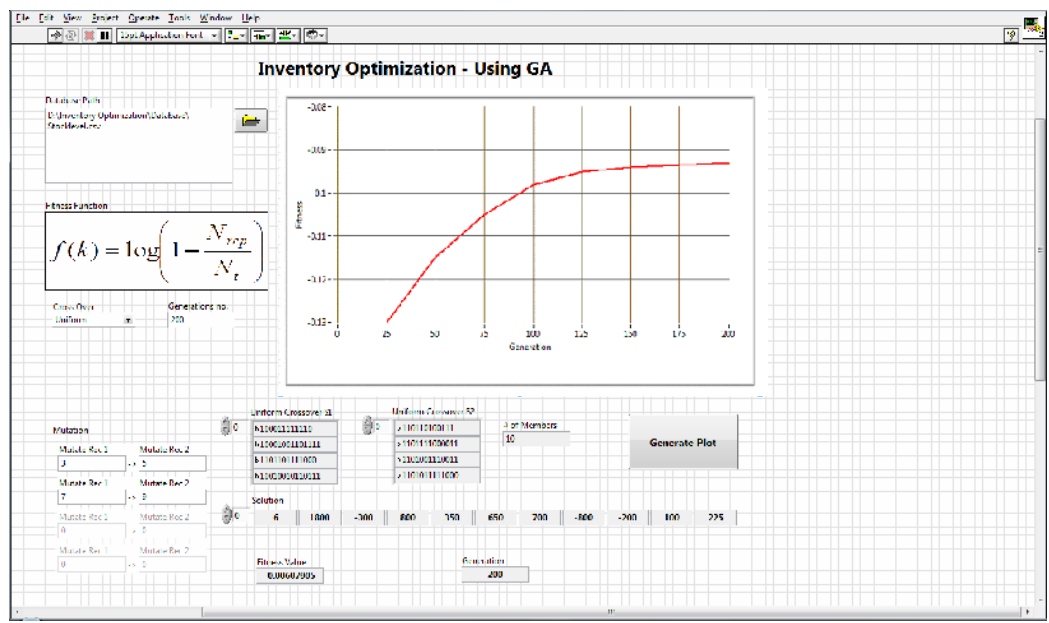

**Fig. 6 Inventory Optimization Tool created in LabVIEW 2009**

In the sample database tabulated in Table 1, the first field comprises of the product Identification (PI) and the other fields are related with the stock levels that were held by the respective ten members of the supply chain network. For example, the first attribute and first field of the database is '7' which refers the Product I.D. '7'. The corresponding fields of the same attribute denote the stock levels of the product I.D. '7' in the respective members of the supply chain. Similarly, different sets of stock levels for different products are held by the database.

As per the proposed analysis based on Uniform Crossover GA, two random initial chromosomes are generated as shown in Fig. 3. These initial chromosomes are subjected for the genetic operators, Uniform Crossover and Mutation. The resultant chromosome thus obtained after the application of crossover and mutation is shown in Fig. 5.

Fig. 6 shows the main window of the tool created for inventory optimization using genetic algorithm in LabVIEW 2009. The tool created is based on the uniform crossover genetic algorithm.

The window displays the fitness function used, uniform crossover sample bit pattern randomly generated with probability of 0.5, the points at which mutation should occur, the end iteration count upon which the fitness function stabilizes, no. of supply chain members and population size.

The best chromosome obtained after the required iterations and the fitness value and plot generated for the iteration value are also displayed.

The organization can decide about the quantum of iterations for running the simulation to arrive at the optimal solution. As long as minimization of the fitness function is still possible, then the iteration continues till such a time that no improvement in the fitness function value is noticeable. After a certain number of iterations, if the fitness function value is not improving from the previous iterations, then this is an indication that the fitness function value is stabilizing and the algorithm has converged towards optimal solution. This inference is useful for deciding the number of iterations for running the GA simulation as well as this may be used as the stopping criteria for the algorithm. For greater accuracy, the number of iterations should be sufficiently increased and run on the most frequently updated large database of past records.

As for our iteration value of '200', the resultant chromosome moved towards the best chromosome after each iterative execution. Hence at the end of the execution of 200th iteration, best chromosome obtained is shown in Fig. 7.

| 6 | 1800 | -300 | 800 | 350 | 650 | 700 | -800 | -200 | 100 | 225 |

Fig. 7 The final best chromosome obtained after 200th iteration

## VI. DISCUSSION OF RESULTS

The final chromosome obtained from the GA based analysis shown in the Fig. 7 is the inventory level that has the potential to cause maximum increase of supply chain cost. It is inferred that controlling this resultant chromosome is sufficient to reduce the loss either due to the holding of excess stocks or due to the shortage of stocks. By focusing on the excess/shortage inventory levels and initiating appropriate steps to eliminate the same at each member of the chain, it is possible to optimize the inventory levels in the upcoming period and thus minimize the supply chain cost.

The organization should take necessary steps to decrease the production of product 6 in the factory by 1800 units to make up for the predicted excess; increase the inventory level of product 1 by 300 units in distribution center 1 to make up for the predicted shortage, reduce inventory level of product 1 by 800 units and 350 units in distribution centers 2 and 3 respectively to make up for the predicted excess.

Agent 1 should decrease the inventory level of product 6 by 650 units. Agent 2 should decrease the inventory level of product 6 by 700 units. Agent 3 and Agent 4 should increase the inventory level of product 6 by 800 units and 200 units respectively to make up for the predicted excess / shortage. The inventory level of product 6 should be decreased by 100 units and 225 units by Agent 5 and Agent 6 respectively. Thus by following the predicted stock levels, the excess/shortage inventory levels can be avoided in the upcoming period and thus the increase of supply chain cost can also be avoided. The analysis extracts an inventory level that made a remarkable contribution towards the increase of supply chain cost, and in turn enabled to predict the future optimal inventory levels to be maintained in all the supply chain members with the aid of these levels. Therefore it is possible to minimize the supply chain cost by maintaining the optimal stock levels that was predicted from the inventory analysis, and thus making the inventory management more effective and efficient.

## VII. CONCLUSION

Inventory management is an important component of supply chain management. An innovative and efficient methodology that uses Genetic Algorithms with Uniform Crossover to precisely determine the most probable excess stock level and shortage level required for inventory optimization in the supply chain such that the total supply chain cost is minimal is proposed using LabVIEW 2009.

The optimized stock level at all members of the supply chain is obtained by following the proposed genetic algorithm. Thus the proposed work gives a better prediction of stock levels amid diverse stock levels at all members of the supply chain. The complexity of increasing the number of products through the supply chain has been resolved by the proposed approach. Products due to which the members of the supply chain incur extra holding or shortage cost are also determined. More specifically, the inventory is optimized in the whole supply chain regardless of the number of products and the number of members in the supply chain. The proposed approach of inventory management has achieved the objectives which are the minimization of total supply chain cost and the determination of the products due to which the supplier endured either additional holding cost or shortage cost.

# REFERENCES

[1]. Sukran Kadipasaoglu, Jennifer Captain and Mark James, "Polymer Supply Chain Management", *International Journal on Logistics Systems and Management*, Vol. 4, No. 2, pp. 233-253, 2008.

[2]. R. King and P. Phumpiu, "Reengineering the food supply chain: The ECR initiative in the grocery industry", *American Journal of Agricultural Economics*, Vol. 78, pp. 1181-1186, 1996.

[3]. Frank Dooley, "Logistics, Inventory Control, and Supply Chain Management", CHOICES: The magazine of food, farm and resource Issues, Vol. 20, No. 4, 4th Quarter 2005.

[4]. R. Wilson, 15th Annual State of Logistics Report. Council of Supply Chain Management Professionals, 2004. Available online: http://www.cscmp.org/.

[5]. R. Ballou, "Business logistics/supply chain management", 5th Ed. Upper Saddle River, NJ: Prentice Hall, 2004.

[6]. M. Fisher, "What is the right supply chain for your product?", Harvard Business Review, Mar/Apr., pp. 105-116, 1997.

[7]. Guangyu Xiong and Hannu Koivisto, "Research on Fuzzy Inventory Control under Supply Chain Management Environment", Lecture Notes in Computer Science, Vol. 2658, pp. 673, 2003.

[8]. M.C. Bonney, "Trends in Inventory Management", *International Journal on Production Economics*, Vol. 35, No. 1-3, pp. 107-114, 1994.

[9]. M. Muller, "Essentials of Inventory Management", NewYork: Amer. Manage. Assoc., 2002.

[10]. S. Nahmias, "Production and Operations Analysis", New York: McGraw-Hill/Irwin, 2004.

[11]. Krishnamurthy, Khorrami and Schoenwald, "Decentralized Inventory Control for Large-Scale Reverse Supply Chains: A Computationally Tractable Approach", *IEEE Transactions on Systems, Man, and Cybernetics*, Vol. 38, No. 4, pp. 551-561, July 2008.

[12]. Kanit Prasertwattana, Yoshiaki Shimizu and Navee Chiadamrong, "Evolutional Optimization on Material Ordering and Inventory Control of Supply Chain through Incentive Scheme", *Journal of Advanced Mechanical Design Systems, and Manufacturing*, Vol. 1, No. 4, pp. 562-573, 2007.

[13]. Mileff, Peter, Nehez, Karoly, "A new inventory control method for supply chain management", 12th International Conference on Machine Design and Production, 2006.

[14]. "Optimization Engine for Inventory Control", Golden Embryo Technologies Pvt. Ltd., Maharastra, India, 2004.

[15]. Jinmei Liu, Hui Gao, Jun Wang, "Air material inventory optimization model based on genetic algorithm", Proceedings of the 3rd World Congress on Intelligent Control and Automation, Vol. 3, pp. 1903 - 1904, 2000.

[16]. C.M. Adams, "Inventory optimization techniques, system vs. item level inventory analysis", 2004 Annual Symposium RAMS - Reliability and Maintainability, pp. 55 - 60, 26-29, Jan, 2004.

[17]. P. Pongcharoen, A. Khadwilard and A. Klakankhai, "Multi-matrix real-coded Genetic Algorithm for minimizing total costs in logistics chain network", *World Academy of Science, Engineering and Technology*, Vol. 26, pp. 458-463, 14-16, December, 2007.

[18]. Greg Scheuffele and Anupam Kulshreshtha, "Inventory Optimization A Necessity Turning to Urgency", SETLabs Briefings, Vol. 5, No. 3, 2007.

[19]. Vecchio and Paschalidis, "Enforcing service-level constraints in supply chains with assembly operations", *Proceedings of IEEE Conference on Decision and Control*, Vol. 5, pp. 5490- 5495, December 2003.

[20]. Ioannis CH. Paschalidis, Yong Liu, Christos G. Cassandras and Christos Panayiotu, "Inventory control for supply chains with service level constraints: A synergy between Large Deviations and Perturbation analysis", *Annals of Operations Research*, Vol. 126, pp. 231-258, 2004.

[21]. J. Sudhir Ryan Daniel and Chandrasekharan Rajendran, "A simulation-based genetic algorithm for inventory optimization in a serial supply chain", *International Transactions in Operational Research*, Vol. 12, pp. 101-127, 2005.

[22]. P. Radhakrishnan, V.M. Prasad and M.R. Gopalan, "Inventory optimization in supply chain management using genetic algorithm", *International Journal of Computer Science and Network Security*, Vol. 9, No. 1, January 2009, pp. 1-8.

[23]. K. Wang and Y. Wang, "Applying Genetic Algorithms to Optimize the Cost of Multiple Sourcing Supply Chain Systems - An Industry Case Study", Studies on Computational Intelligence, Vol. 92, pp. 355-372, 2008.

[24]. Chih-Yao Lo, "Advance of Dynamic Production-Inventory Strategy for Multiple Policies Using Genetic Algorithm", *Information Technology Journal*, Vol. 7, pp. 647-653, 2008.

[25]. Ulrich W. Thonemann and James R. Bradley, "The effect of product variety on supply-chain performance", *European Journal of Operational Research*, Vol. 143 No.3, pp. 548-69, 2002.

[26]. S. Buffett and N. Scott, "An Algorithm for Procurement in Supply Chain Management", *AAMAS-04 Workshop on Trading Agent Design and Analysis*, New York, 2004.

[27]. S. Behzadi, Ali A. Alesheikh and E. Poorazizi, "Developing a Genetic Algorithm to solve Shortest Path Problem on a Raster Data Model", *Journal on Applied Sciences*, Vol. 8, No. 18, pp. 3289-3293, 2008.

[28]. Aphirak Khadwilard and Pupong Pongcharoen, "Application of Genetic Algorithm for Trajectory Planning of Two Degrees of Freedom Robot Arm With Two Dimensions", *Thammasat International Journal on Science and Technology*, Vol. 12, No. 2, April- June 2007.

[29]. M.A. Sharbafi, M. Shakiba Herfeh, Caro Lucas and A. Mohammadi Nejad, "An Innovative Fuzzy Decision Making Based Genetic Algorithm", *World Academy of Science, Engineering and Technology*, Vol. 19, pp. 172-175, May 2006.

[30]. Thomas Butter, Franz Rothlauf, Jorn Grahl, Hildenbrand Jens Arndt, "Developing Genetic Algorithms and Mixed Integer Linear Programs for Finding Optimal Strategies for a Student's Sports Activity", *Working Papers in Information Systems, University of Mannheim*, 2006.

[31]. S.A. Qureshi, S.M. Mirza and M. Arif, "Fitness Function Evaluation for Image Reconstruction using Binary Genetic Algorithm for Parallel Ray Transmission Tomography", *International Conference on Emerging Technologies*, 2006. ICET '06. 13-14, Nov. 2006, pp. 196-201.

[32]. Saifuddin Md. Tareeq, Rubayat Parveen, Liton Jude Rozario and Md. Al-Amin Bhuiyan, "Robust Face detection using Genetic Algorithm", *Journal on Information Technology*, Vol. 6, No. 1, pp. 142-147, 2007.

[33]. M. Soryani and N. Rafat, "Application of Genetic Algorithms to Feature Subset Selection in a Farsi OCR", *World Academy of Science, Engineering and Technology*, Vol. 18, pp. 113-116..

[34]. William M. Spears and K.A. De Jong, "On the Virtues of Uniform Crossover", 4th International Conference on Genetic Algorithms, La Jolla, California, July 1991.

[35]. Syswerda, Gilbert, "Uniform Crossover in Genetic Algorithms", Proc. 3rd International Conference on Genetic Algorithms, Morgan Kaufman Publishing, 1989.

[36]. Riccardo Poli and W.B. Langdon, "On the Search Properties of Different Crossover Operators in Genetic Programming", Proceedings of Genetic Programming'98, Madison, Wisconsin, 1998.

## ABOUT AUTHORS

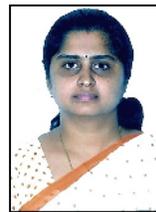

**Mrs. S. Narmadha** is working as Assistant Professor in the Department of Computer Science and Engineering in Park College of Engineering and Technology, Coimbatore. She obtained her Bachelor's degree in Computer Science and Engineering from Tamilnadu College of Engineering, Coimbatore under Bharathiar University and Master's degree in Mechatronics from Vellore Institute of Technology, Vellore. She is currently pursuing Ph.D. under Anna University, Chennai. She has 8 years of Teaching Experience and 2 years of Industrial experience. She has published 14 papers in International Conferences, 2 papers in International journals and a book on 'Open Source Systems'. She is life member of ISTE. Her field of interest includes Supply Chain Management, Automation, Database Management Systems, Virtual Instrumentation, Soft Computing Techniques and Image Processing.

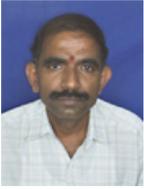

**Dr. V. Selladurai** is the Professor and Head, Department of Mechanical Engineering and Principal, Coimbatore Institute of Technology, Coimbatore, India. He holds a Bachelor's degree in Production Engineering, a Master's degree in Industrial Engineering specialisation and a Ph.D. degree in Mechanical Engineering. He has two years of industrial experience and 22 years of teaching experience. He has published over 90 papers in the proceedings of the leading National and International Conferences. He has published over 35 papers in international journals and 22 papers in national journals. His areas of interest include Operation Research, Artificial Intelligence, Optimization Techniques, Non-Traditional Optimization Techniques, Production Engineering, Industrial and Manufacturing Systems, Industrial Dynamics, System Simulation, CAD/CAM, FMS, CIM, Quality Engineering and Team Engineering.

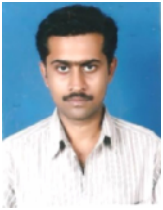

**Mr. G. Sathish** is a full time research scholar under Anna University, Coimbatore. He holds a Bachelor's degree and a Master's degree in Computer Science and Engineering. He has 8 years of industrial experience. He has published 10 papers in the proceedings of the leading International Conferences. His field of interest includes Supply Chain Management, Optimization Techniques, Data Mining, Knowledge Discovery, Automation, Soft Computing Techniques and Image Processing.